\documentclass[conference]{IEEEtran}
\IEEEoverridecommandlockouts
\usepackage{cite}
\usepackage{amsmath,amssymb,amsfonts}
\usepackage{algorithmic}
\usepackage{graphicx}
\usepackage{textcomp}
\usepackage{xcolor}

\def\BibTeX{{\rm B\kern-.05em{\sc i\kern-.025em b}\kern-.08em
    T\kern-.1667em\lower.7ex\hbox{E}\kern-.125emX}}
\begin{document}

\title{Deep Learning Powered Estimate of The Extrinsic Parameters on Unmanned Surface Vehicles\\
}

\author{\IEEEauthorblockN{1\textsuperscript{st} Yi Shen}
\IEEEauthorblockA{\textit{University of Michigan Ann Arbor} \\
Ann Arbor, MI, USA\\
shenrsc@umich.edu}
\and

\IEEEauthorblockN{1\textsuperscript{st} Hao Liu}
\IEEEauthorblockA{\textit{Northeastern University} \\
Shenyang, China\\
liuhao@stumail.neu.edu.cn}
\and

\IEEEauthorblockN{2\textsuperscript{nd} Chang Zhou}
\IEEEauthorblockA{\textit{Columbia University} \\
USA\\
mmchang042929@gmail.com}
\and

\IEEEauthorblockN{3\textsuperscript{rd} Wentao Wang}
\IEEEauthorblockA{\textit{University of Southern California} \\
LA, CA, USA\\
wwang047@usc.edu}
\and

\IEEEauthorblockN{4\textsuperscript{th} Zijun Gao}
\IEEEauthorblockA{\textit{Northeastern University} \\
USA\\
gao.zij@northeastern.edu}
\and

\IEEEauthorblockN{5\textsuperscript{th} Qi Wang}
\IEEEauthorblockA{\textit{Northeastern University} \\
USA\\
bjwq2019@gmail.com}
}

\maketitle

\begin{abstract}
Unmanned Surface Vehicles (USVs) are pivotal in marine exploration, but their sensors' accuracy is compromised by the dynamic marine environment. Traditional calibration methods fall short in these conditions. This paper introduces a deep learning architecture that predicts changes in the USV’s dynamic metacenter and refines sensors' extrinsic parameters in real time using a Time-Sequence General Regression Neural Network (GRNN) with Euler angles as input. Simulation data from Unity3D ensures robust training and testing. Experimental results show that the Time-Sequence GRNN achieves the lowest mean squared error (MSE) loss, outperforming traditional neural networks. This method significantly enhances sensor calibration for USVs, promising improved data accuracy in challenging maritime conditions. Future work will refine the network and validate results with real-world data.
\label{sec:abstract}
\end{abstract}


\section{Introduction}
The rapid advancement of autonomous maritime vehicles, particularly USVs, has significantly broadened the scope of marine exploration and utilization. As USVs become more prevalent, there is a pressing need to equip these vehicles with reliable navigation and mapping systems, leading to a demand for high-accuracy sensors.

However, the deployment of sensors on USVs is fraught with challenges primarily due to the dynamic nature of marine environments. The motion of the vessel, notably the dynamic metacenter, critically impacts the sensors' extrinsic parameters—their orientation and position relative to the USV. Traditional calibration methods often fail under such fluctuating conditions, leading to inaccuracies in data that can jeopardize the USV’s operational integrity.

Recent advances in deep learning have shown great promise in addressing similar challenges in autonomous systems. Deep neural networks (DNNs) and convolutional neural networks (CNNs) have dramatically enhanced capabilities in various domains, including autonomous navigation \cite{deng2023long, deng2023plgslam}, game strategy \cite{zhang2024development}. Moreover, the use of deep learning models for real-time sensor calibration in dynamic environments has shown significant promise. Yan et al. \cite{10.1007/978-981-15-3753-0_34} demonstrated the effectiveness of machine learning models in calibrating and adjusting the parameters of autonomous vehicle sensors in real time, enhancing their adaptability to dynamic conditions.

In industrial applications, neural networks facilitate significant improvements in object recognition, scene understanding, and decision-making processes. For example, the integration of deep learning techniques in 3D reconstruction and industrial automation has been explored by Tong et al. \cite{10458957, 9723528}. Additionally, Ding et al. \cite{8665155} investigated the use of multiple monocular cameras for vehicle pose and shape estimation, highlighting the potential of deep learning to enhance depth and pose estimation capabilities. Yan et al. \cite{10.1007/978-981-15-3753-0_34} utilized LSTM neural networks to predict oil production, showing the versatility of deep learning in handling time-series data. Zhang and Makris \cite{9358255} applied machine learning to detect Spectre attacks, emphasizing the broad applicability of these techniques beyond physical sensors.

Advances in reinforcement learning have also contributed significantly, as demonstrated by Li et al. \cite{re_learning}, who developed a reinforcement learning-based routing algorithm for large street networks. Furthermore, research by Chen et al. \cite{chen2021multimodal} on semi-supervised learning for 3D objects showcases the continuous evolution of deep learning models in understanding and interpreting complex environments. The study by Weng and Wu \cite{Weng2024} on cyber security indexes and data protection measures further exemplifies the diverse applications of deep learning techniques in various fields.

In addition, the application of LSTM neural networks in predicting lightning location and movement has also demonstrated the versatility of these models in handling complex environmental data \cite{8936293}. Research by Castillo et al. on enhanced image capture using computer-vision networks has opened new avenues for improving sensor accuracy and reliability \cite{Castillo21}.

Similarly, the use of contextual hourglass networks for semantic segmentation of high-resolution aerial imagery has advanced the field of environmental monitoring and disaster response \cite{Li19}. The integration of deep learning techniques for anomaly detection in time series data further underscores the potential of these models to handle diverse data types and applications \cite{lymperopoulos2022exploiting}.

Fuzzy control systems have also shown considerable potential in dealing with uncertainties in nonlinear systems. Merazka et al.\cite{7958730} presented a fuzzy state-feedback control method for uncertain nonlinear MIMO systems, demonstrating robustness against uncertainties and enhancing system stability. Similarly, Zhang and Yang\cite{8105904} proposed an observer-based fuzzy adaptive sensor fault compensation method for uncertain nonlinear strict-feedback systems, which improves tracking performance and fault tolerance. Another notable contribution by Merazka et al.\cite{7958728} involves high-gain observer-based adaptive fuzzy control for multivariable nonlinear systems, which enhances control performance by addressing uncertainties effectively.

Inspired by these advancements, this article proposes a novel network architecture leveraging Radial Basis Function (RBF) networks to predict changes in the dynamic metacenter of the vessel and adjust the sensors' extrinsic parameters in real time. This approach not only enhances the accuracy of the sensors' data but also contributes significantly to the broader field of autonomous navigation in challenging marine environments.

\section{Related Work}
\label{sec:related_work}
\subsection{Metacenter}
\begin{figure}[htbp]
    \centering
    \includegraphics[width=0.5\linewidth]
    {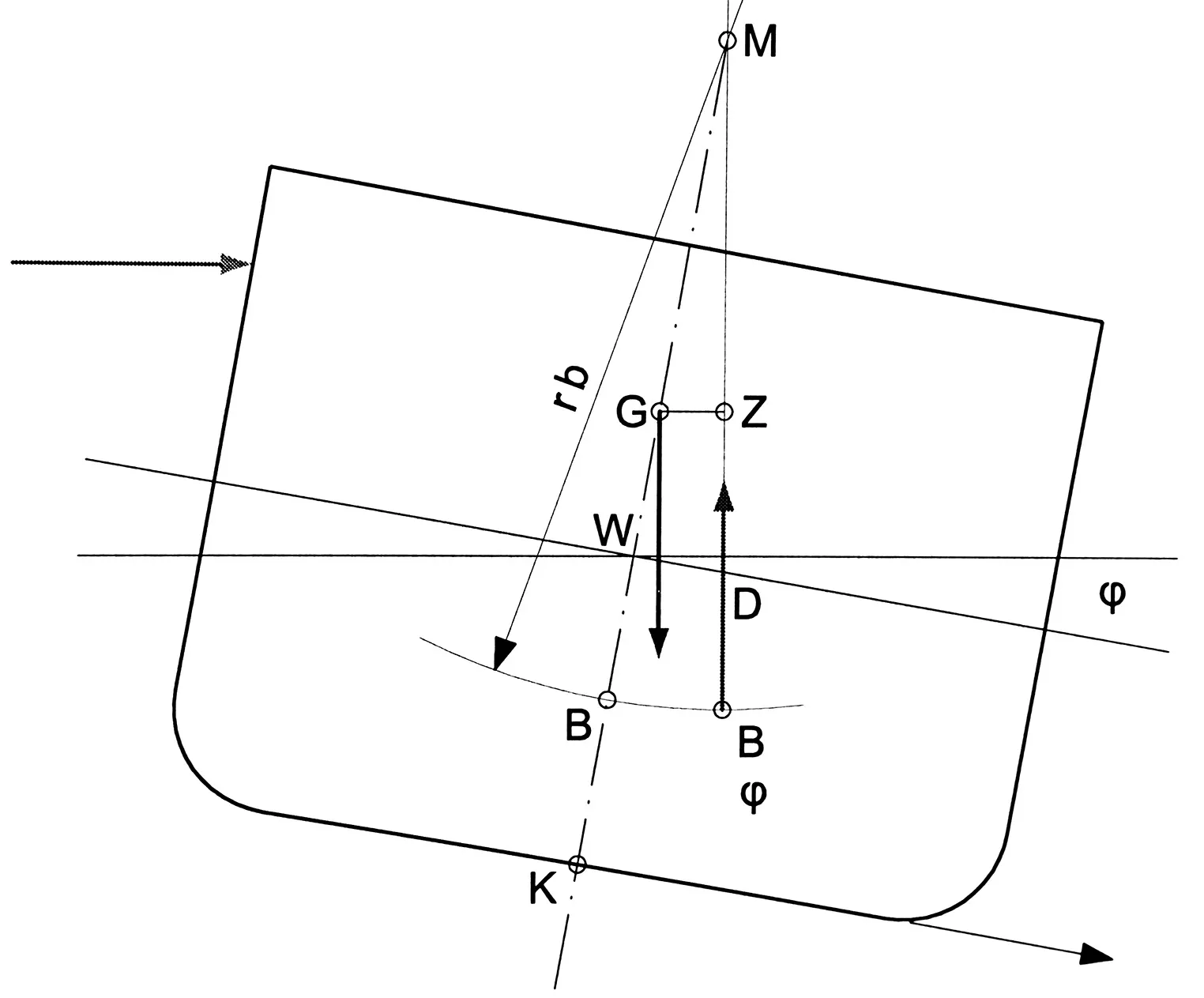}\\
    \caption{Ship stability condition.\cite{vasilescu2020influence}}
    \label{fig:metacenter}
\end{figure}

\
The metacenter is an important concept in the field of fluid mechanics, particularly in the study of the stability of floating bodies such as ships and boats. As shown in Fig.~\ref{fig:metacenter}, $G,M,B$ are the gravity center, the metacenter, and the buoyant center separately. The metacenter refers to the point where the buoyant force, which acts vertically upwards through the center of buoyancy of a partially submerged object, is considered to act when the object is tilted or displaced from its vertical orientation. Since the floating body tilts around it, it is the precise center of the ship's frame. However, The location of the metacenter is dependent on the shape of the waterline and the distribution of buoyancy, both of which change as a boat tilts and the metacenter's location is dynamically changing.

\subsection{Extrinsic Parameters Estimate on USVs}\
Extrinsic parameters describe the position and orientation of a sensor such as a Lidar relative to a reference frame(In the case of USVs, usually the ship frame). Unlike robots on the ground where the center of the robot frame is fixed, as mentioned before, the metacenter is the precise center of the ship's frame system but it is moving. As a result, the exact extrinsic parameters are changing all the time. Then how to estimate the extrinsic parameters properly is an important problem unsolved.

The position of the dynamic metacenter is a critical factor in calibrating the sensors on USVs. This position can be determined using the Euler angles (roll, pitch, and yaw) as inputs shown in equation~\ref{eular_to_pos}.
\begin{equation}
    \label{eular_to_pos}
    s = [x,y,z] = \mathbf{f}(\theta, \phi, \psi)
\end{equation}

Let $\Theta$ represent the roll angle, $\Phi$ represent the pitch angle, and $\psi$ represent the yaw angle. The position of the metacenter $s$ in Cartesian coordinates $[x,y,z]$ can be expressed as a function of these angles. Here, $\mathbf{f}$ is a complex function that maps the Euler angles to the metacenter coordinates. This function accounts for the non-linear dynamics of the vessel's motion. $\mathbf{f}$ currently has no analytic solution but the numerical solution can be calculated by the finite element model(FEM). However, the calculation of FEM for one position can take a few minutes especially when the resolution is large, making it impossible to satisfy the real-time requirement. Leveraging the advantage of the neural network(NN) to Approximate arbitrary complex functions, the framework proposed in this article attempts to predict the position of the metacenter of the USV in real-time based on the Euler angle which can be achieved easily by IMU on the USV.

\section{Method and Implementation}
\label{sec:method}
Recently, thanks to its ability to approximate complex functions, deep learning, and NN have become popular. This article employs several NN structures to predict the position of the metacenter of the USV.

\subsection{Radial basis function network}
\begin{figure}[htbp]
    \centering
    \includegraphics[width=0.5\linewidth]
    {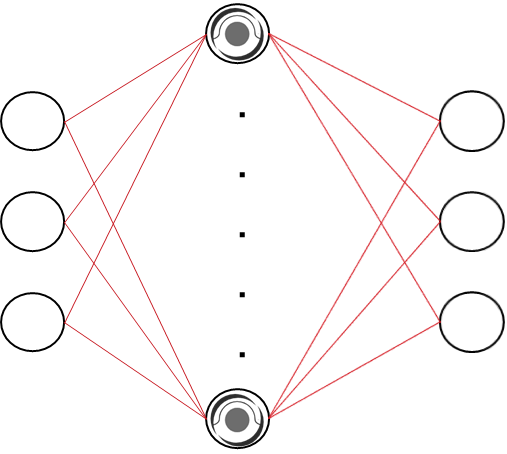}\\
    \caption{The structure of the RBF.}
    \label{fig:RBF}
\end{figure}

The radial basis function (RBF) network uses radial basis functions as activation functions. These functions are typically Gaussian functions, which are defined based on the distance between the input vector and the center of each neuron as shown in equation~\ref{Guassian}. During the training, not only the weights between each connected neuron are trained, but the centers and spreads of the radial basis functions can also be adjusted. The adjustment on radial basis functions changes the basis of the hidden layer and reflects input into a proper high-dimension space where classification or regression is easier than in a low-dimension space. As a result, it is supposed to have better performance than a normal fully connected NN. The structure of a typical RBF is built with an input layer, a hidden layer, and an output layer which is shown in Fig.~\ref{fig:RBF}.

\subsubsection{Gaussian kernal function}
\begin{equation}
\label{Guassian}
    \Phi_{i}(||\mathbf{x}-\mathbf{c}_{i}||) = exp(-||\frac{\mathbf{x}-\mathbf{c}_{i}}{\mathbf{\sigma}_i}||^2)
\end{equation}
Let $x$ be the input vector, $c_i$ be the center of the $i-th$ neuron, and $\sigma_i$ be the spread (or width) of the Gaussian function for the $i-th$ neuron. In the equation, $\Phi_i$ is the activation function for the $i-th$ neuron.
$||x-c_i||$ represents the Euclidean distance between the input vector $x$ and the center $c_i$ of the $i-th$ neuron. $\sigma_i$  controls the width of the Gaussian function and determines how far the influence of the center $c_i$ extends. The exponential function ensures that the activation decreases rapidly as the distance between $x$ and 
$c_i$ increases. $\mathbf{\sigma}$ and $\mathbf{c}$ can be learned or calculated directly from data. In this article, those parameters are learned.

\subsection{General regression neural network}
\begin{figure}[htbp]
    \centering
    \includegraphics[width=0.5\linewidth]
    {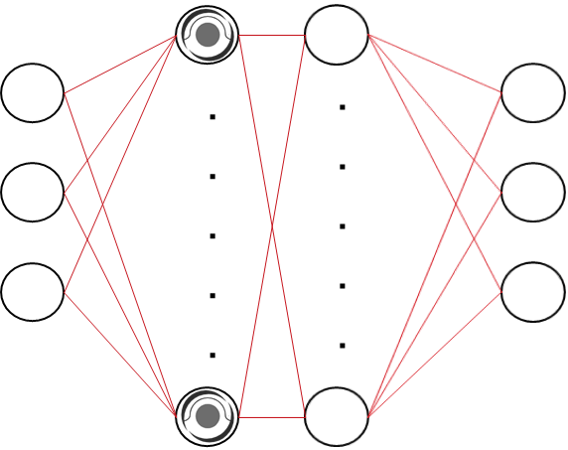}\\
    \caption{The structure of the GRNN.}
    \label{fig:GRNN}
\end{figure}

A common improved network based on RBF is the general regression neural network(GRNN). GRNN is built on the basis of nonparametric regression, which essentially uses the data from the sample as a post hoc, and the output of the network is calculated based on the principle of maximum probability. With the addition of a hidden layer, GRNN is particularly suitable for RBF networks based on its good nonlinear approximation ability.
With the addition of a hidden layer, it is especially suitable for curve-fitting problems based on the good nonlinear approximation ability of RBF networks. The additional hidden layer is a summarization layer and the structure is shown in Fig.~\ref{fig:GRNN}.

\subsection{Time sequence information}
\begin{figure}[htbp]
    \centering
    \includegraphics[width=0.7\linewidth]
    {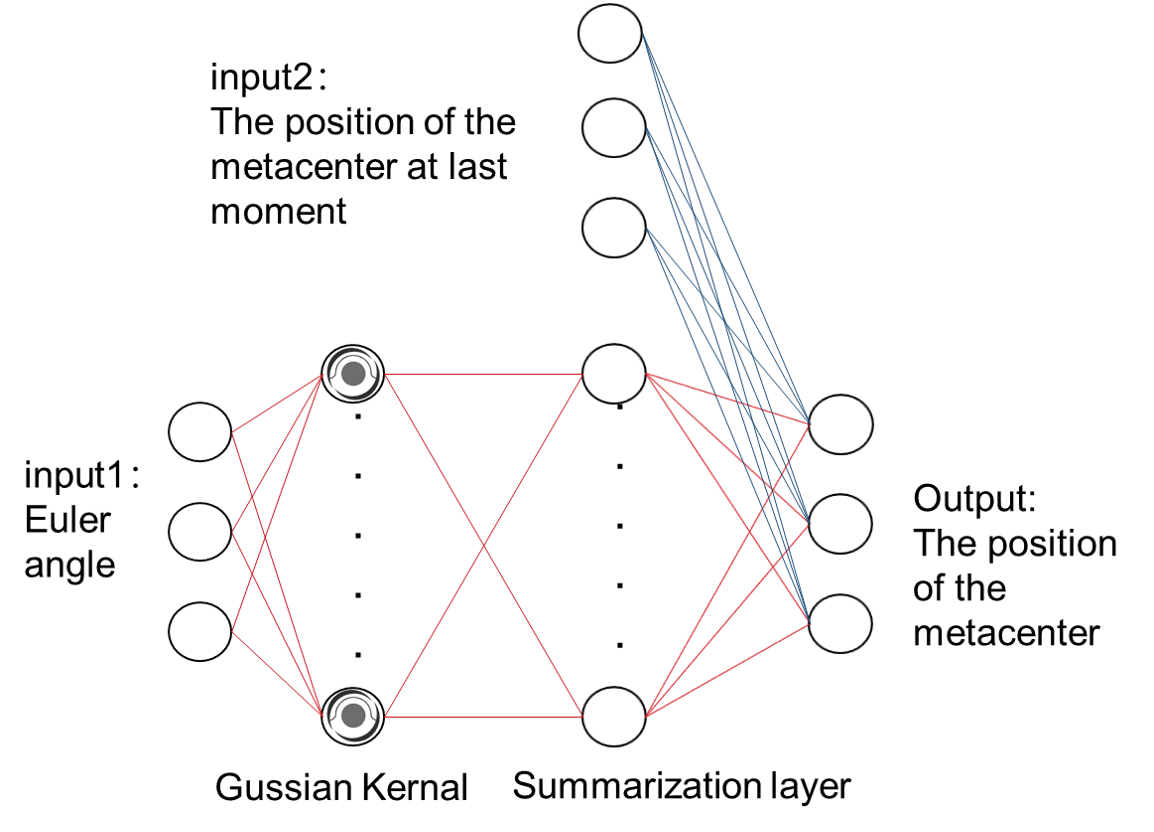}\\
    \caption{The structure of the time sequence information network.}
    \label{fig:time_sequence}
\end{figure}

In terms of previous networks in our task, the output is the position of the metacenter and the input is the Euler angle. However, the position of the metacenter is dynamic and related to time. Thus, time can be another useful piece of information that has not been leveraged by previous network structures. Directly using time as input will introduce problems but it can be exploited indirectly shown in Fig.~\ref{fig:time_sequence}. This structure is based on GRNN with an extra input as predicted metacenter at the last moment.

\subsection{Loss function}
The mean square error(MSE) or L2 loss function is chosen. It takes the average of the squared difference between the predictions and the target values so that it fits well with the errors in the Cartesian coordinate.

\section{Experiments and results}
\label{sec:result}
A series of simulation experiments in Unit3D are conducted to validate the efficiency of the time-sequence GRNN. Unit3D generates Euler angles and the metacenter as the benchmark. Then metacenter positions predicted by different methods during the simulation are used to compare the error.

\subsection{Origin of data}
Unity3D is a game development tool developed by Unity Technologies. In Unity3D, a model of USV can be built based on the known design parameters. Then the model can be simulated in environments powered by the physical engine(PhysX) and the position of the metacenter can be accessed. The simulation speed in Unity3D is much faster than normal FEM software such as Abaqus and a lot of useful simulation data can be accessed quickly. 

\subsection{Data collection and process}
Data was collected at a sampling rate of 10 Hz over multiple trials. The raw Euler angles were preprocessed to reduce the high-frequency characteristics of the data using a 3D Gaussian filter. Besides, a variance filter is applied to filter out those strange data points to achieve better performance. The processed data was then fed into the Time-Sequence GRNN and other control groups to predict the metacenter position in real time.

\subsection{Results}
During the training process, we use a batch size of 128 and perform 64 updates using the Adam optimizer. Besides, the init step size for Adam is set to $1e^{-2}$. We trained the same data with 4 networks and the Table\ref{comparison_table} reports the average loss in the training set, and the testing set.

\begin{table}[h!]
    \begin{center}
        \caption{\small{Performances of different network structures}}
        \begin{tabular}{ l c c } 
        \hline
        \textbf{Network structure} & \textbf{Loss in training set} & \textbf{Loss in testing set} \\
         \hline
         \hline
         Fully connected network & 17.53 & 131.49   \\ 
         \hline
         RBF & 6.67 & 15.55  \\ 
         \hline
         GRNN & 6.41 & 14.33 \\ 
         \hline
         Time-sequence GRNN & 3.10 & 6.65  \\ 
         \hline
        \end{tabular}
        \label{comparison_table}
    \end{center}
\end{table}
\subsubsection{Fully connected network}
This network has 2 hidden layers with 20 Relu neurons on each layer.
\subsubsection{RBF}
The structure of the RBF is the same as in Fig.~\ref{fig:RBF} with 20 neurons in the hidden layer.
\subsubsection{GRNN}
The structure of the GRNN is the same as in Fig.~\ref{fig:GRNN} with 20 neurons in each hidden layer.
\subsubsection{Time-sequence GRNN}
The same as GRNN except the time sequence information is applied.


\begin{figure}[htbp]
    \centering
    \includegraphics[width=0.625\linewidth]
    {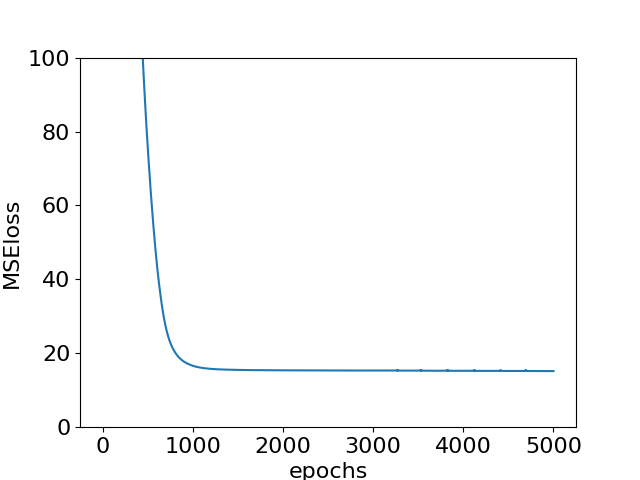}\\
    \caption{MSE loss of the fully connected network}
    \label{fig:connected_network_train}
\end{figure}

\begin{figure}[htbp]
    \centering
    \includegraphics[width=0.625\linewidth]
    {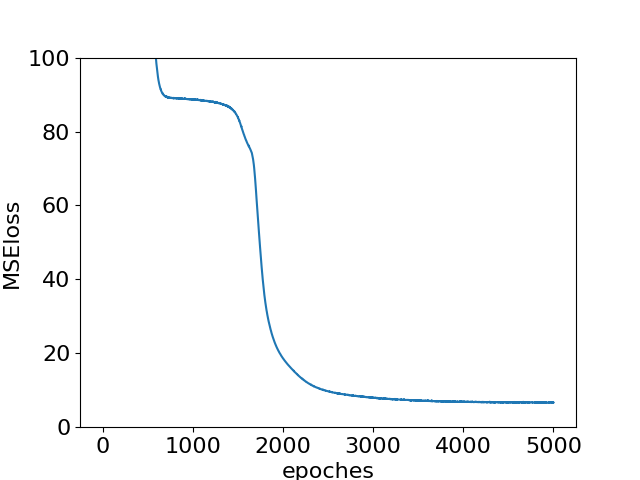}\\
    \caption{MSE loss of the RBF}
    \label{fig:RBF_train}
\end{figure}

\begin{figure}[htbp]
    \centering
    \includegraphics[width=0.625\linewidth]
    {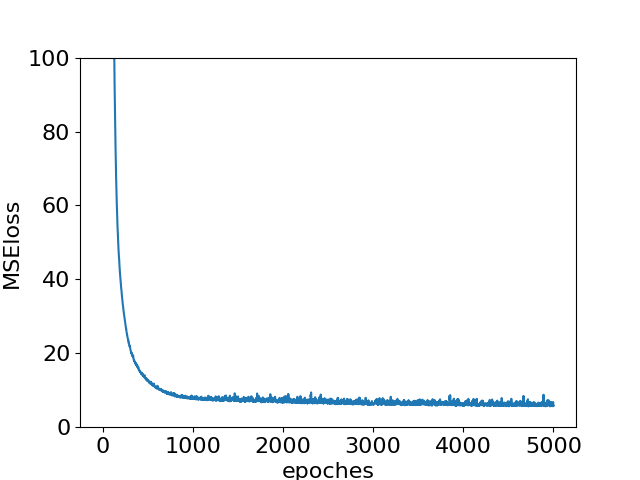}\\
    \caption{MSE loss of the GRNN}
    \label{fig:GRNN_train}
\end{figure}

\begin{figure}[htbp]
    \centering
    \includegraphics[width=0.625\linewidth]
    {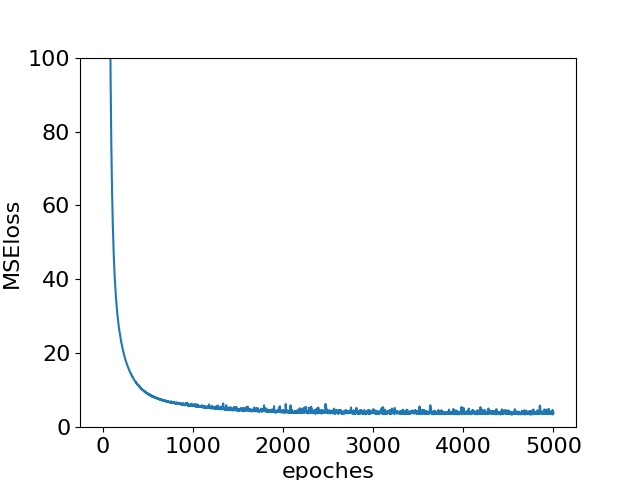}\\
    \caption{MSE loss of the time sequence GRNN}
    \label{fig:time_sequence_GRNN_train}
\end{figure}

To visualize the data, Fig.~\ref{fig:connected_network_train}, Fig.~\ref{fig:RBF_train}, Fig.~\ref{fig:GRNN_train} and Fig.~\ref{fig:time_sequence_GRNN_train} report the MSE Loss over the training process. All training will stop at 5000 iterations.

\section{Discussion}
\subsection{Consideration of the network}
The performance of a network is affected by several parameters. Firstly, hyperparameters such as the learning rate, optimizer, and iteration times can have a meaningful impact on the behavior of our system. $1e^{-2}$ which is a relatively large learning rate is chosen so that the model can converge fast. To avoid diverging risk introduced by a large learning rate, the Adams optimizer is applied since it can adapt the learning rate. 

Another important issue is the initialization of the values of neurons in the network. All zero or equal-value initialization will cause the symmetry problem and lead to the no-learning problem. Normal initialization is used here and the result is positive. However, due to the gradient explosion and the gradient vanishing problem, this method is not suited for those networks with a large depth. In this case, Xavier should be chosen. Xavier has been tried but the result is not as good as normal initialization in our scope and was thus abandoned.

The training set and the test set both come from the Unity3D simulation but no real data from a real USV are tested. It is unknown if the network can predict the position of the metacenter precisely and get the extrinsic parameters correctly under real working conditions.

\subsection{Analysis of result}
The training and testing loss data visualized in Fig.~\ref{fig:connected_network_train},~\ref{fig:RBF_train},~\ref{fig:GRNN_train},~\ref{fig:time_sequence_GRNN_train} and quantified in Table~\ref{comparison_table}
illustrate the effectiveness of four neural network architectures: the Fully Connected Network, RBF Network, GRNN, and Time-Sequence GRNN.

The Time-Sequence GRNN shows superior performance with the lowest MSE losses (3.10 in training and 6.65 in testing). This testing error corresponds to an average error of 2.58 cm, which is precise given the USV's 9.5m length. By leveraging sequential data, the Time-Sequence GRNN provides continuous, context-aware predictions, with loss plots showing a rapid, steady decrease and stabilization at a lower value compared to other networks. This efficiency and consistency are crucial for dynamic environments like USV sensor calibration.

\subsection{Advantage of the work}
The Time-Sequence GRNN demonstrates excellent generalization with minimal disparity between training and testing losses, indicating reduced overfitting. Its robust performance, highlighted by minimal fluctuations in MSE loss, stems from its ability to incorporate feedback from sequential data points.

In conclusion, the Time-Sequence GRNN sets a new benchmark in loss metrics and showcases a forward-thinking approach to neural network design for dynamic systems. Its effective use of temporal data makes it promising for real-time analysis and response in evolving conditions.

\subsection{Limitation of the work}
Though the performance of the time sequence network is good, accumulated errors may be introduced since the position of the metacenter at the previous moment is used as input. The error can be magnified as time passes by and the magnitude of the influence is currently unknown.

During the experiment, all parameters are tuned and chosen based on the performance of the basic fully connected network first and then applied to RBF, GRNN, and the Time-Sequence GRNN. Therefore, the performance of different network structures can be compared evenly. However, the parameters achieved in this way may not fully exploit the power of the Time-Sequence GRNN.

\subsection{Future work}
Based on the discussion above, 2 major areas can be improved in the future for this project.
\begin{itemize}
    \item Data in the real world should be recorded and introduced to test the proposed method, especially those boundary cases that might not be covered by the simulation data. A solid conclusion can be made only if excellent performance on real-world data has been achieved.
    \item Since the structure of the time sequence network shows its bright potential, it should be leveraged to develop more powerful structures. The depth, width, and layers can be modified and the newly created network should be trained and tested to get better performance.
\end{itemize}

\section{Conclusion}
This study proposed the Time-Sequence GRNN as a powerful tool for predicting extrinsic parameters under the influence of a dynamic metacenter. After evaluating the performance of different neural network architectures, the superior capability of the Time-Sequence GRNN in handling dynamic systems is proved. By incorporating temporal dynamics, the Time-Sequence GRNN demonstrated not only the lowest MSE losses but also remarkable consistency and generalization, with minimal fluctuations across training and testing phases. These attributes make it particularly suited for applications like sensor calibration in dynamic environments such as USVs. Future research should focus on validating these findings in real-world conditions and further exploring the potential of time-sequence models to enhance predictive accuracy and robustness in various other dynamic systems.

\bibliographystyle{IEEEtran}
\bibliography{ref.bib}

\end{document}